\documentclass[journal,twoside,web]{ieeecolor}
\usepackage{generic}
\usepackage{cite}
\usepackage{amsmath,amssymb,amsfonts}
\usepackage{graphicx}
\usepackage{algorithm}
\usepackage{algorithmic}
\usepackage{multirow}
\usepackage{orcidlink}
\usepackage{booktabs} 
\usepackage{stfloats} 
\usepackage{threeparttable} 

\usepackage{textcomp}
\def\BibTeX{{\rm B\kern-.05em{\sc i\kern-.025em b}\kern-.08em
    T\kern-.1667em\lower.7ex\hbox{E}\kern-.125emX}}

\usepackage[T1]{fontenc}
\usepackage[utf8]{inputenc}

\usepackage{hyperref}
\hypersetup{
    unicode=true,
    colorlinks=true,
    linkcolor=blue,
    citecolor=blue,
    urlcolor=blue,
    filecolor=blue,
    pdfborder={0 0 0},
    hypertexnames=false
}

\begin{document}
\title{BioLLMAgent: A Hybrid Framework with Enhanced Structural Interpretability for Simulating Human Decision-Making in Computational Psychiatry}
\author{Fei Zuo$^{~\orcidlink{0009-0007-9493-0581}}$, Kezhi Wang$^{~\orcidlink{0000-0001-8602-0800}}$, \IEEEmembership{Senior Member, IEEE}, Xiaomin Chen$^{~\orcidlink{0000-0001-9267-355X}}$, Yizhou Huang$^{~\orcidlink{0009-0008-2189-2470}}$
\thanks{This work was partly supported by Horizon Europe HarmonicAI project under grant number 101131117, and UKRI grant EP/Y03743X/1. K. Wang
would like to acknowledge the support in part by the Royal Society
Industry Fellowship (IF/R2/23200104). Corresponding author: Kezhi Wang, Yizhou Huang.}
\thanks{Fei Zuo is with the School of Computer Science and Technology, East China Normal University, 3663 North Zhongshan Road, Shanghai 200062, China (e-mail: 71255901041@stu.ecnu.edu.cn). }
\thanks{Xiaomin Chen is with the Department of Computer Science, University of Reading, RG6 6UR, UK (e-mail: xiaomin.chen@reading.ac.uk).}
\thanks{Kezhi Wang and Yizhou Huang are with the Department of Computer Science,  Brunel University of London, UB8 3PH, UK (email: kezhi.wang@brunel.ac.uk; yizhou.huang2@brunel.ac.uk).}}

\maketitle

\begin{abstract}
Computational psychiatry aims to transform mental health research through mathematical models, yet existing approaches face a fundamental trade-off: traditional reinforcement learning (RL) models offer interpretability but lack behavioral realism, while Large Language Model (LLM) agents generate realistic behaviors but lack structural interpretability for scientific analysis. In response, we introduce \texttt{BioLLMAgent}, a novel hybrid framework that combines the interpretability of computational models with the behavioral generation capabilities of LLMs to create a structurally interpretable simulation tool for psychiatric research. The framework consists of three core components: (i) an Internal RL Engine implementing validated cognitive models (e.g., Outcome-Representation Learning) that simulates experience-driven value learning; (ii) an External LLM Shell that captures high-level cognitive strategies and therapeutic interventions through natural language prompts; and (iii) a Decision Fusion Mechanism that integrates both components through weighted averaging of utility values. Through comprehensive experiments on the Iowa Gambling Task across six datasets spanning healthy controls and addiction populations, we demonstrate that \texttt{BioLLMAgent} accurately reproduces human behavioral patterns, maintains excellent parameter identifiability with core cognitive parameters showing correlations $>0.67$, exhibits controllable responses to prompt manipulations in large models, and explores in silico encoding of cognitive behavioral therapy principles with behaviorally plausible effects. Multi-agent social dynamics simulations generate hypothesis-generating findings that community-wide educational interventions may outperform targeted individual treatments. \texttt{BioLLMAgent} provides a powerful computational sandbox for testing theoretical hypotheses and intervention strategies, offering new opportunities for mechanistic understanding of decision-making deficits in computational psychiatry. The framework has been validated on two tasks within the decision-making domain: reward-punishment learning (Iowa Gambling Task) and temporal discounting (Delay Discounting task), demonstrating cross-task generalization of core components. Generalization to other cognitive domains (working memory, social cognition, attention control) remains to be demonstrated.
\end{abstract}


\begin{figure*}[!tbp]
\centering
\includegraphics[width=0.85\textwidth]{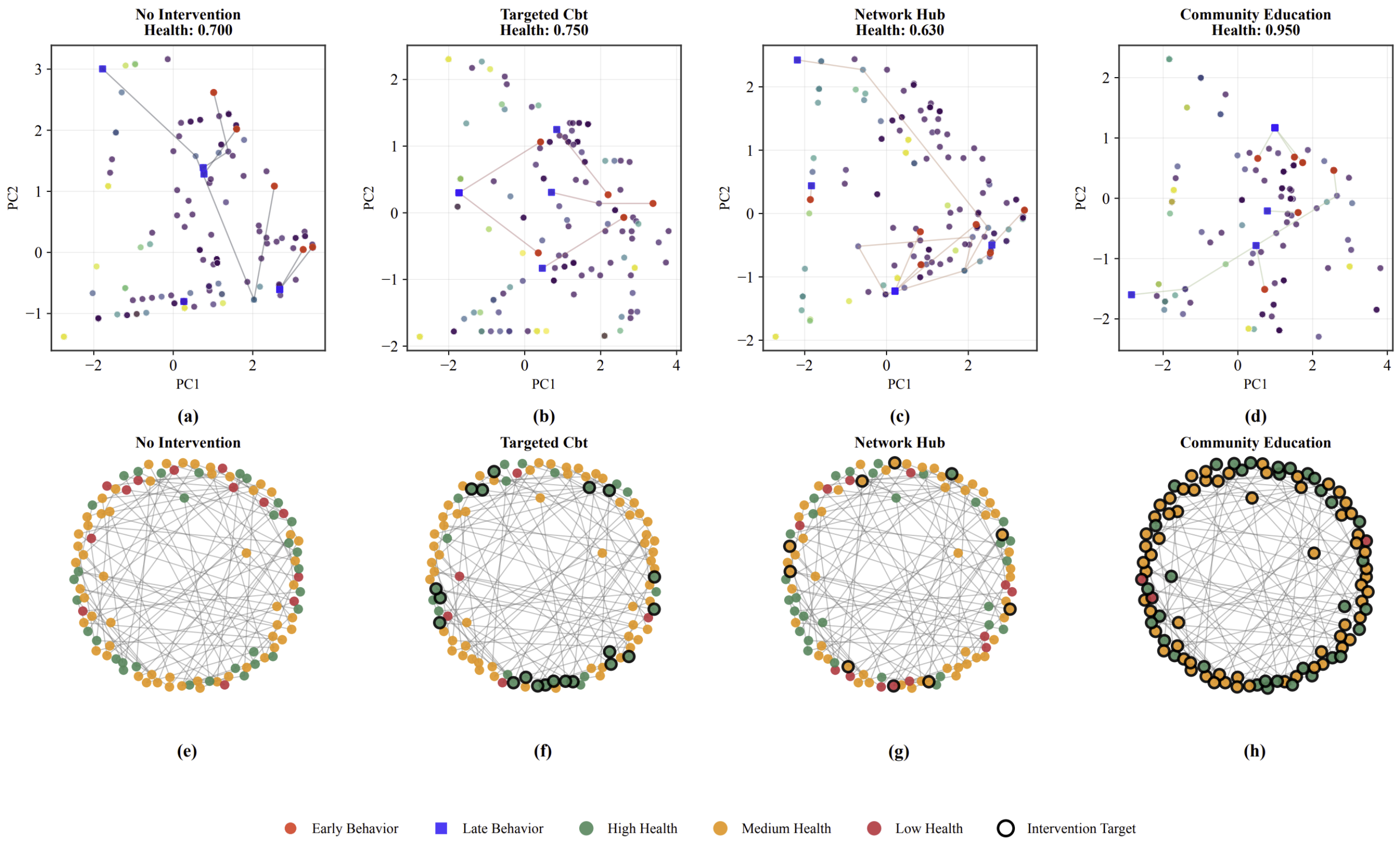}
\caption{Network-level intervention effects. Top: PCA visualization of behavioral states under (a) No Intervention (0.700), (b) Targeted CBT (0.750), (c) Network Hub (0.630), (d) Community Education (0.950). Bottom: Network structures with health levels. Community Education achieves highest performance with cohesive behavioral convergence and uniform network-wide improvements.}
\label{fig:network_pca_analysis}
\end{figure*}

\section{Introduction}
\label{sec:introduction}
\IEEEPARstart{H}{uman} decision-making, particularly under uncertainty and reward-punishment contingencies, has been a central focus of computational psychiatry. AI methods have shown broad success in biomedicine, from molecular modeling \cite{yue2024integration} to medical image diagnosis \cite{xu2024deep}, and are now increasingly applied to mental health through computational psychiatry, which promises to redefine our understanding of mental disorders, moving beyond descriptive labels to quantifiable, mechanistic models of cognition \cite{khaleghi_computational_2022,adams_computational_2015}. Reinforcement learning (RL) has been central to this effort, with models like Prospect Valence Learning (PVL) and Outcome-Representation Learning (ORL) being applied to tasks like the Iowa Gambling Task (IGT) \cite{steingroever_comparison_2013,ahn_decision-making_2014}. These models offer mathematical interpretability, linking cognitive deficits in conditions like addiction to specific parameters, like insensitivity to losses or hypersensitivity to rewards \cite{janbesaraei_igt_2024}. However, this approach faces a persistent "model war," where no single framework consistently captures the full spectrum of human decision-making \cite{janbesaraei_human_2025}. While mathematically elegant, these models often fail to generate behavior reflecting variability and contextual nuances, creating a significant gap between theoretical insight and behavioral realism \cite{steingroever_performance_2013}.

In parallel, the rise of Large Language Models (LLMs) has introduced a new paradigm for behavioral simulation \cite{park_generative_2023}. LLM-based generative agents can plan, reason, and interact in simulated environments with remarkable authenticity \cite{11091354}. This has spurred interest in creating "virtual patients" for clinical training and AI assessment, where agents can simulate specific disease symptoms through fluid, context-aware conversations \cite{wang_patient-_2024,yuan_improving_2025}. These agents seem to perfectly address the behavioral generation shortcomings of traditional computational models. However, their application as rigorous scientific tools is hindered by a fundamental lack of structural interpretability \cite{lu_prompting_2025}. An agent's choices emerge from billions of opaque parameters, ungrounded in any explicit psychological or neuroscientific theory. This "black-box" nature makes it impossible to determine precisely why a decision was made, limiting their utility for mechanistic discovery \cite{rmus_generating_2025}. Thus, computational psychiatry faces a trade-off: RL models offer interpretability without realism, while LLMs offer realism without interpretability.

To bridge this gap, we introduce \texttt{BioLLMAgent}, a novel hybrid framework that integrates the strengths of both approaches to achieve structural interpretability alongside behavioral realism. Our core innovation is to embed a validated, interpretable computational model as an \textbf{Internal RL Engine} within an \textbf{External LLM Shell}. This design explicitly models two distinct cognitive drivers: an \textit{Internal Drive}, generated by the RL engine, which simulates experience-based value learning from direct environmental interaction (representing slow, trial-and-error habit formation); and an \textit{External Drive}, generated by the LLM, which captures high-level beliefs, situational reasoning, or external instructions (such as a therapist's advice) to form decision priors. By constraining the LLM's contribution to specific, theoretically meaningful cognitive functions, this dual-component architecture makes the agent's decision process both analyzable and realistic.

This architecture enables quantitative predictions from the mathematical model and qualitative reasoning from the LLM to mutually validate and constrain one another. We validated \texttt{BioLLMAgent} through experiments using the IGT across diverse clinical and healthy populations. Beyond individual-level validation, we demonstrate the framework's potential for agent-based simulation through large-scale network simulations (Fig.~\ref{fig:network_pca_analysis}), where community-wide educational interventions achieve the highest average health score (0.950), significantly outperforming individual treatments (0.750). This work offers a path toward accelerating research in computational psychiatry, particularly for decision-making and impulsivity-related paradigms. The framework has been validated on two canonical tasks (Iowa Gambling Task, Delay Discounting) that assess complementary aspects of decision deficits in addiction: reward-punishment learning and temporal impulsivity. Cross-task validation (Fig.~\ref{fig:delay_discounting_gen}) demonstrates successful generalization of modular architecture (seamless RL engine substitution: ORL $\rightarrow$ Hyperbolic Discounting), fusion mechanism (consistent $\omega$ effects), and LLM-based priors (task-specific prompt engineering). Extension to other cognitive domains represents an important future direction.

\section{Related Work}
\label{sec:related_work}

\subsection{Computational Psychiatry and Reinforcement Learning Models}
Computational psychiatry uses mathematical frameworks to define mental disorders as disruptions in cognitive algorithms \cite{khaleghi_computational_2022,adams_computational_2015}. Reinforcement learning (RL) models have been central to this effort, particularly for decision-making tasks like the Iowa Gambling Task (IGT) \cite{steingroever_performance_2013,ahn_decision-making_2014}. Models such as Prospect Valence Learning (PVL), which captures reward sensitivity \cite{fridberg_cognitive_2010}, Value-plus-Perseverance (VPP), which incorporates habits \cite{ligneul_sequential_2019}, and the Outcome-Representation Learning (ORL) model used in our framework \cite{noauthor_outcomerepresentation_nodate}, offer interpretability. Their parameters can serve as cognitive biomarkers, identifying patterns like altered loss aversion in addiction \cite{ahn_decision-making_2014,groman_reinforcement_2022}. Despite their success, these models are limited by their abstraction, failing to capture contextual and narrative elements of human choice. This has led to "model wars" where no approach prevails, and an inability to generate realistic "digital subjects" for simulation-based research \cite{steingroever_comparison_2013,janbesaraei_human_2025}.

\subsection{Large Language Models as Behavioral Agents}
Large language models (LLMs) have recently revolutionized behavior generation, with agents capable of simulating nuanced human interactions in virtual environments \cite{park_generative_2023}. Using natural language prompts, these agents can role-play complex scenarios, leading to applications such as creating "virtual patients" for clinical training that replicate symptoms of depression or addiction \cite{wang_patient-_2024,yuan_improving_2025}. Their strength is behavioral realism; they can approximate human performance in decision-making tasks like the IGT, handle ambiguity, and generate diverse, context-aware outcomes \cite{park_simulating_nodate,zeng_dynamic_nodate}. However, this realism comes at the cost of interpretability. The "black-box" nature of LLMs, where decisions arise from billions of opaque parameters, prevents their use for rigorous scientific inquiry, as they are not grounded in validated cognitive theories \cite{lu_prompting_2025,rmus_generating_2025}. Furthermore, their controllability is inconsistent, especially in smaller models, making it difficult to systematically simulate interventions like therapy \cite{cook_virtual_2025}.

\begin{figure*}[!t]
\centering
\includegraphics[width=1\textwidth]{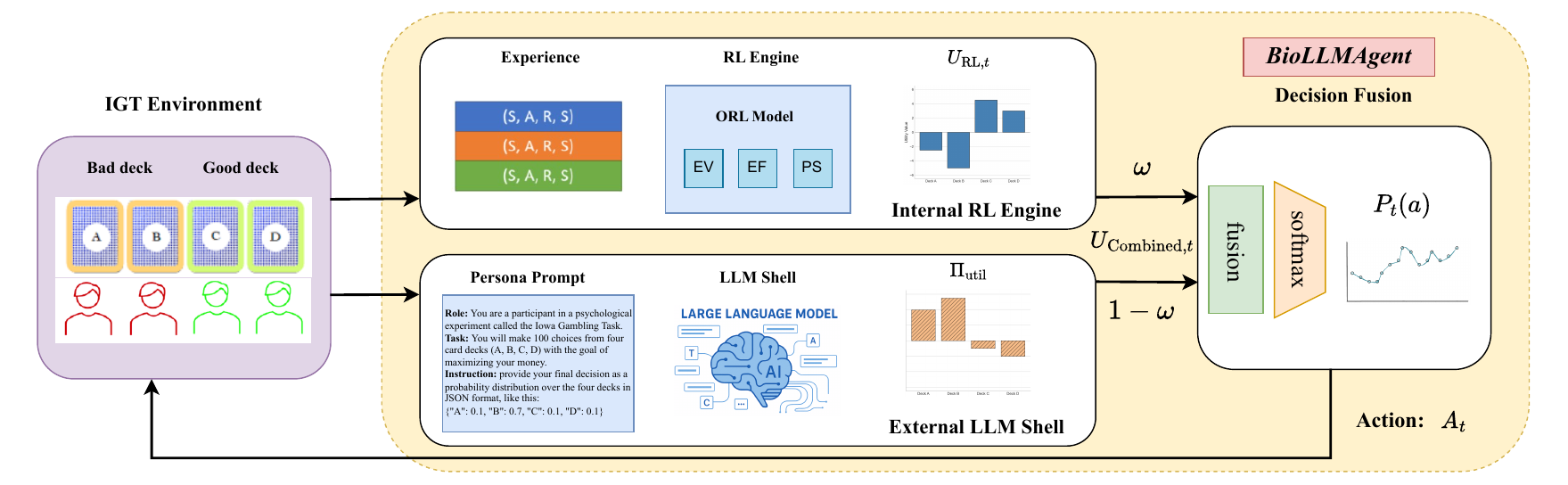}
\caption{BioLLMAgent framework architecture integrating IGT Environment, Internal RL Engine (ORL model), External LLM Shell, and Decision Fusion mechanism. Parameter $\omega$ controls RL-LLM balance. The Internal RL Engine processes experience to generate Expected Value (EV), Expected Frequency (EF), and Perseveration (PS) utilities, while the External LLM Shell uses persona prompts to simulate complete IGT trials, generating probability distributions that are averaged and converted to static utility-scale priors $\Pi_{\text{util}}$.}
\label{fig:framework_schematic}
\end{figure*}

\subsection{Hybrid RL-LLM Approaches}
To combine the strengths of both paradigms, hybrid RL-LLM approaches have emerged, aiming to pair RL's structured learning with the reasoning of LLMs. In fields like robotics, this often involves using an LLM for high-level planning and RL for low-level motor control \cite{noauthor_saycan_nodate}. While some work has explored using hybrids for social simulation in cognitive science \cite{rmus_generating_2025,jiang_cognitive_2025}, their application to computational psychiatry remains limited. Critically, existing hybrid models have not been designed with structural interpretability as a primary goal. They often lack a clear separation of components for analysis and have not integrated clinically-validated RL models (like ORL) as an internal cognitive engine. Moreover, they are rarely validated against clinical datasets with rigorous parameter recovery, and their potential for large-scale social and network-level intervention simulation remains unexplored. Our work with \texttt{BioLLMAgent} directly addresses these gaps by proposing a modular hybrid framework that preserves the interpretability of a validated RL model while leveraging an LLM to inject behavioral realism and enable controllable, large-scale simulations for psychiatric research.

\section{Methodology}
In this paper, we introduce an innovative hybrid agent framework named \texttt{BioLLMAgent}, designed to merge the interpretability of computational Reinforcement Learning (RL) models with the behavioral generation capabilities of Large Language Models (LLMs) to create a structurally interpretable simulation tool for psychiatric research. As illustrated in Fig.~\ref{fig:framework_schematic}, the framework operates through a clear information flow: starting from the IGT Environment, proceeding through parallel processing in the Internal RL Engine and External LLM Shell, and culminating in the Decision Fusion mechanism. This modular design decomposes the agent's decision-making process into independently analyzable endogenous and exogenous driving components through three core modules: (i) an \textbf{Internal RL Engine} implementing the ORL model for experience-driven value learning (endogenous drive); (ii) an \textbf{External LLM Shell} capturing high-level cognitive strategies through persona prompts (exogenous drive); and (iii) a \textbf{Decision Fusion Mechanism} that integrates their utility outputs to produce interpretable and realistic decision sequences.

\subsection{Task Paradigm: The Iowa Gambling Task (IGT)}
All our computer simulations are based on the classic Iowa Gambling Task (IGT), a widely used experimental paradigm for assessing experience-driven decision-making abilities, particularly in clinical populations with decision-making deficits \cite{bechara1994,steingroever_performance_2013,ahn_decision-making_2014}. The IGT has been extensively validated in computational psychiatry research for modeling addiction-related decision-making deficits \cite{fridberg_cognitive_2010,janbesaraei_igt_2024,groman_model-free_2019}.

In this task, participants are required to make repeated selections from four virtual decks of cards (A, B, C, D) over 100 trials, with the goal of maximizing their total cumulative earnings \cite{steingroever_data_2015}. The reward and loss structure of these four decks is carefully designed to create a conflict between short-term high returns and long-term net gains \cite{steingroever_comparison_2013}:
\begin{itemize}
    \item \textbf{Disadvantageous Decks (Decks A \& B):} These two decks offer higher immediate monetary rewards but are accompanied by larger or more frequent penalties, leading to a net loss if chosen over the long term.
    \item \textbf{Advantageous Decks (Decks C \& D):} These two decks provide lower immediate rewards, but their penalties are significantly smaller and less frequent, thus ensuring a long-term net gain.
\end{itemize}
Furthermore, the task introduces an asymmetry in the dimension of outcome frequency: Decks A and C have a higher frequency of losses (50\%), whereas Decks B and D have a lower frequency of losses (10\%) \cite{yechiam_using_2005,ligneul_sequential_2019}. This complex design makes the IGT an ideal testbed for investigating how decision-makers trade off between multiple conflicting dimensions. To ensure computational rigor, we formalize the task as a single-state Markov Decision Process (MDP), where the action space comprises the four decks and the reward function is stochastic according to the chosen deck's payoff schedule \cite{daw_model-based_2011,chen_model-based_2021}.

\subsection{The BioLLMAgent Framework}
The core architecture of \texttt{BioLLMAgent} lies in its modular design, which provides a detailed mathematical formalization of the framework's components.

\begin{algorithm}[!t]
 \caption{Bayesian Inference \& Simulation}
 \label{alg:main_algorithm}
\begin{algorithmic}[1]
 \renewcommand{\algorithmicrequire}{\textbf{Input:}}
 \renewcommand{\algorithmicensure}{\textbf{Output:}}
 \REQUIRE Human data $D_{\text{human}}$
 \ENSURE Posterior samples $\Theta$
 \STATE \textbf{Phase 1: External Prior Generation}
 \STATE \textsc{GetState}() \COMMENT{Simulate IGT}
 \STATE $\pi_t \leftarrow \pi_{\text{LLM}}(a|s_t, P)$ \COMMENT{Query LLM}
 \STATE $\Pi_{\text{trials}}$.add($\pi_t$)
 \STATE $\Pi_{\text{prob}} \leftarrow \frac{1}{T} \sum_{\pi \in \Pi_{\text{trials}}} \pi$ \COMMENT{Average}
 \STATE $\Pi_{\text{util}} \leftarrow \textsc{Convert}(\Pi_{\text{prob}})$ \COMMENT{Scale}
 \STATE \textbf{return} $\Pi_{\text{util}}$
 \STATE \textbf{Phase 2: Bayesian Inference}
 \STATE $\Theta = \{\theta^{(i)}\}_{i=1}^N$ of model $M$
 \STATE $\mathcal{L}(D_{\text{human}}|\Theta) \leftarrow \dots$ \COMMENT{Likelihood}
 \STATE Sample from $P(\Theta|D_{\text{human}})$ to get $\{\Theta_s\}_{s=1}^S$
 \STATE \textbf{Phase 3: Simulation}
 \FOR{each $\Theta_s$, each subject $i=1$ to $N$}
    \STATE $\text{state}_1^{(i)} \leftarrow M.\textsc{Init}(\theta_s^{(i)})$
    \FOR{trial $t=1$ to $T_i$}
        \STATE $U_{M,t}^{(i)} \leftarrow M.\textsc{GetUtility}(\text{state}_t^{(i)})$
        \STATE $U_{\text{Comb},t}^{(i)} \leftarrow (1-w) \cdot U_{M,t}^{(i)} + w \cdot \Pi_{\text{util}}$
        \STATE $a_{\text{sim},t}^{(i)} \sim \text{Categorical}(\text{Softmax}(U_{\text{Comb},t}^{(i)}))$
        \STATE $\text{state}_{t+1}^{(i)} \leftarrow M.\textsc{Update}(\dots)$
    \ENDFOR
 \ENDFOR
 \STATE \textbf{return} $P(\Theta|D_{\text{human}}), D_{\text{sim}}$
\end{algorithmic}
\end{algorithm}

\subsubsection{Module 1: The Internal RL Engine}
The Internal RL Engine, positioned at the center of the framework architecture, is designed to simulate trial-and-error based value learning driven by direct experience with the IGT environment \cite{groman_reinforcement_2022,yang_forgetting_2025}. This module is "plug-and-play," allowing for the integration of various computational models \cite{ahn_computational_2016}. The modular design ensures framework generalizability beyond ORL/IGT. Any RL model must implement a standardized interface: $\textsc{GetUtility}(\text{state}) \rightarrow \mathbf{U}_{RL}$, enabling seamless model substitution for different domains (e.g., hyperbolic discounting for temporal choice, Fig.~\ref{fig:delay_discounting_gen}). The fusion mechanism operates independently, requiring only utility vectors as input. Extension to new tasks involves: (1) selecting validated domain-appropriate RL models, (2) implementing $\textsc{GetUtility}$, (3) re-validating parameter identifiability. ORL was chosen for its strong IGT validation record. In the current work, we employ the \textbf{Outcome-Representation Learning (ORL)} model as the core engine \cite{haines2018,noauthor_outcomerepresentation_nodate}. The ORL model was chosen because it explicitly separates the learning processes for expected value (EV), expected frequency (EF), and perseveration (PS) components, as visualized in Fig.~\ref{fig:framework_schematic}, providing us with a more refined, multi-component characterization of the decision-making process \cite{janbesaraei_human_2025}.

As detailed in \textbf{Phase 2} of Algorithm~\ref{alg:main_algorithm}, the internal RL engine parameters are estimated through Bayesian inference on human behavioral data. The algorithm initializes the parameter space $\Theta = \{\theta^{(i)}\}_{i=1}^N$ for the ORL model and uses MCMC sampling to obtain posterior distributions $P(\Theta|D_{\text{human}})$ (Steps 9-11). This ensures that the endogenous cognitive components remain grounded in empirical evidence while maintaining full parameter interpretability.

The ORL model consists of several main parts:
\begin{enumerate}
    \item \textbf{Asymmetric Learning of Value and Frequency:} The model does not use a subjective utility function but learns directly from the objective gain/loss outcomes $X_t$ \cite{steingroever_validating_2013}. It maintains and updates expected value $EV_j(t)$ and expected frequency $EF_j(t)$ (for deck $j$ at trial $t$) separately, using two independent learning rates—a reward learning rate $A_{rew}$ and a punishment learning rate $A_{pun}$—to capture differences in sensitivity to gains and losses \cite{groman_model-free_2019,chen_model-based_2021}.
    
    \begin{itemize}
        \item \textbf{Expected Value Update:} The expected value is updated according to Equation~\eqref{eq:1}:
        \begin{equation}
        \label{eq:1}
        \begin{array}{l}
            EV_j(t+1) = EV_j(t) \\
            \qquad + 
            \begin{cases}
                A_{rew} \cdot (X_t - EV_j(t)) & \text{if } X_t \ge 0 \\
                A_{pun} \cdot (X_t - EV_j(t)) & \text{if } X_t < 0
            \end{cases}
        \end{array}.
        \end{equation}
      
        \item \textbf{Expected Frequency Update:} Similarly, the expected frequency is updated as shown in Equation~\eqref{eq:2}:
        \begin{equation}
        \label{eq:2}
        \begin{array}{l}
            EF_j(t+1) = EF_j(t) \\
            \qquad + 
            \begin{cases}
                A_{rew} \cdot (\text{sgn}(X_t) - EF_j(t)) & \text{if } X_t \ge 0 \\
                A_{pun} \cdot (\text{sgn}(X_t) - EF_j(t)) & \text{if } X_t < 0
            \end{cases}
        \end{array},
        \end{equation}
        where the $\text{sgn}(X_t)$ function returns 1, 0, and -1 for positive, zero, and negative outcomes, respectively.
    \end{itemize}
    \item \textbf{Perseveration Process:} The model also includes an independent perseveration weight $PS_j(t)$ to capture a value-independent tendency to repeat a choice (perseveration) or to switch (exploration) \cite{ligneul_sequential_2019,yang_forgetting_2025}. After each trial, the perseveration weight for the chosen deck $j$ is reset to 1, while the weights for all decks decay according to a forgetting/decay parameter $K$.
    \item \textbf{Decision Utility Integration:} At the time of decision, the final integrated utility of an action $a$ for the internal engine, $U_{\text{RL}, t}(a)$, is computed as a linear weighted sum of the three components mentioned above, as formalized in Equation~\eqref{eq:3}:
    \begin{equation}
    \label{eq:3}
    U_{\text{RL}, t}(a) = EV_t(a) + \beta_F \cdot EF_t(a) + \beta_P \cdot PS_t(a),
    \end{equation}
    where $\beta_F$ and $\beta_P$ are the weight parameters for frequency preference and perseveration preference, respectively. Thus, the ORL model contains a total of five core free parameters: $\{A_{rew}, A_{pun}, K, \beta_F, \beta_P\}$ \cite{mkrtchian_reliability_nodate}.
\end{enumerate}

\begin{table}[!tbp]
\caption{ORL Model Parameters and Their Cognitive Interpretations}
\label{tab:orl_params}
\centering
\footnotesize
\begin{tabular}{llp{0.42\columnwidth}}
\toprule
\textbf{Param.} & \textbf{Range} & \textbf{Cognitive Interpretation} \\
\midrule
$A_{rew}$ & [0, 1] & Reward learning rate \\
$A_{pun}$ & [0, 1] & Punishment learning rate \\
$K$ & [0, 5] & Forgetting/perseveration decay \\
$\beta_F$ & $\mathbb{R}$ & Frequency weight (wins vs.\ losses) \\
$\beta_P$ & $\mathbb{R}$ & Perseveration weight \\
$\theta$ & [0, $\infty$) & Inverse temperature (exploration) \\
$\omega$ & [0, 1] & Fusion weight (external vs.\ internal) \\
\bottomrule
\end{tabular}

\vspace{4pt}
\begin{tablenotes}
\footnotesize
\item Note: $A_{rew}$--$\beta_P$ are core RL Engine parameters; $\theta$ controls choice noise; $\omega$ balances internal and external drives.
\end{tablenotes}
\end{table}

\subsubsection{Module 2: The External LLM Shell}
The External LLM Shell, depicted in the upper right of Fig.~\ref{fig:framework_schematic}, formalizes external instructions or high-level cognitive beliefs as a static cognitive prior through persona prompts. This module represents specific cognitive biases or therapeutic guidance \cite{park_generative_2023,xie_can_2024}. The cognitive prior is extracted from a Large Language Model (LLM) through a three-step process, designed to convert its dynamic, natural language-based reasoning capabilities into a stable utility vector suitable for mathematical computation and fusion with the RL engine outputs \cite{rmus_generating_2025}:

The extraction process is formalized in \textbf{Phase 1} of Algorithm~\ref{alg:main_algorithm}, which addresses the fundamental challenge of converting dynamic, token-based LLM outputs into static utility vectors compatible with mathematical decision models. The algorithm simulates the complete IGT task using a "pure LLM agent" endowed with a specific persona (Steps 2-3), collecting trial-specific probability distributions $\pi_{t, \text{LLM}}(a|s_t, P)$ across all trials (Step 4).

\begin{enumerate}
    \item \textbf{Policy Sampling:} First, we run a "pure LLM agent" endowed with a specific persona, allowing it to complete the entire IGT task \cite{lu_prompting_2025}. At each trial $t$, we record the action probability distribution $\pi_{t, \text{LLM}}(a)$ output by the LLM for all actions $a \in A$, as implemented in Steps 2-4 of Algorithm~\ref{alg:main_algorithm}.

    \item \textbf{Static Prior Aggregation:} To obtain a prior that represents the stable decision-making tendency of this persona throughout the task, we average the probability distributions from all $T$ trials to get a single, static probability vector $\Pi_{\text{prob}}$, as defined in Equation~\eqref{eq:4} and implemented in Step 5 of Algorithm~\ref{alg:main_algorithm}:
    \begin{equation}
    \label{eq:4}
    \Pi_{\text{prob}} = \frac{1}{T} \sum_{t=1}^{T} \pi_{t, \text{LLM}}.
    \end{equation}
    This vector represents the overall preference distribution of the LLM persona for the four options.

    \item \textbf{Conversion to Utility Scale:} Since the probability scale (range [0, 1]) is not compatible with the utility scale (arbitrary real numbers) computed by the internal RL model, we need to convert the probability prior to the utility scale. We achieve this by centering and scaling it to obtain a utility vector $\Pi_{\text{util}}$, as shown in Equation~\eqref{eq:5} and Step 6 of Algorithm~\ref{alg:main_algorithm}, where the sign of its elements represents a preference for or aversion to an option relative to the average:
    \begin{equation}
    \label{eq:5}
    \Pi_{\text{util}} = \text{scale}(\Pi_{\text{prob}} - \bar{\Pi}_{\text{prob}}).
    \end{equation}
    This transformed static vector $\Pi_{\text{util}}$ serves as the effective exogenous drive fused with dynamic RL utilities. The static prior is a deliberate simplifying assumption justified on three grounds: (1) \textbf{Cognitive validity:} it models trait-level stable dispositions (personality, internalized beliefs, therapeutic guidance) consistent with dual-process theories; (2) \textbf{Methodological priority:} parameter identifiability and causal interpretability outweigh capturing all cognitive dynamics, cleanly separating ``internal learning'' (RL) from ``external influence'' (LLM); (3) \textbf{Pragmatic scope:} proof-of-concept for hybrid architecture, with dynamic extensions as future work.
\end{enumerate}

\begin{table}[!tbp]
\caption{Breakdown of Datasets Used in the Current Study}
\label{tab:datasets}
\centering
\begin{tabular}{lcc}
\toprule
\textbf{Dataset} & \textbf{N} & \textbf{Population} \\
\midrule
Ahn et al. (HC) \cite{ahn_decision-making_2014} & 48 & Healthy Controls \\
Horstmann et al. \cite{steingroever_data_2015} & 162 & Healthy Controls \\
Kjome et al. \cite{kjome_relationship_2010} & 19 & Healthy Controls \\
Maia \& McClelland \cite{maia_reexamination_2004} & 40 & Healthy Controls \\
Ahn et al. (Amphetamine) \cite{ahn_decision-making_2014} & 38 & Amphetamine Users \\
Ahn et al. (Heroin) \cite{ahn_decision-making_2014} & 43 & Heroin Users \\
\bottomrule
\end{tabular}

\vspace{10pt}
\begin{tablenotes}
\footnotesize
\item Note: HC = Healthy Controls. All datasets used the modified Iowa Gambling Task with 100 trials per participant.
\end{tablenotes}
\end{table}

\begin{figure*}[!tbp]
\centering
\includegraphics[width=0.85\textwidth]{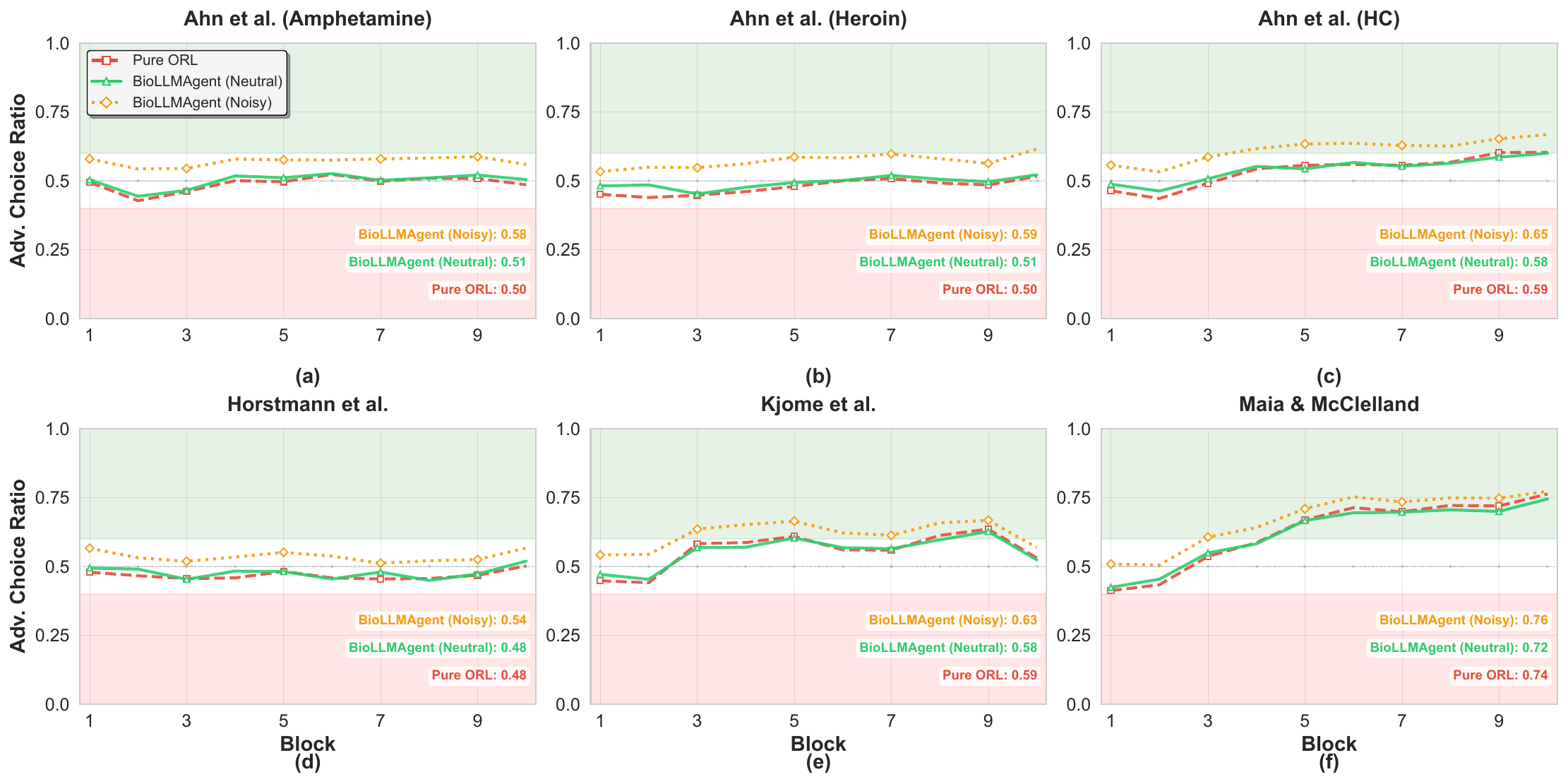}
\caption{Behavioral trajectory validation using GPT-4o. Choice patterns for (a) Amphetamine, (b) Heroin, (c) Healthy controls, (d-f) Additional datasets. Lines: Pure ORL (red), BioLLMAgent neutral (green), noisy control (orange), human data (black).}
\label{fig:traj_gpt4o}
\end{figure*}

\begin{figure*}[!tbp]
\centering
\includegraphics[width=0.85\textwidth]{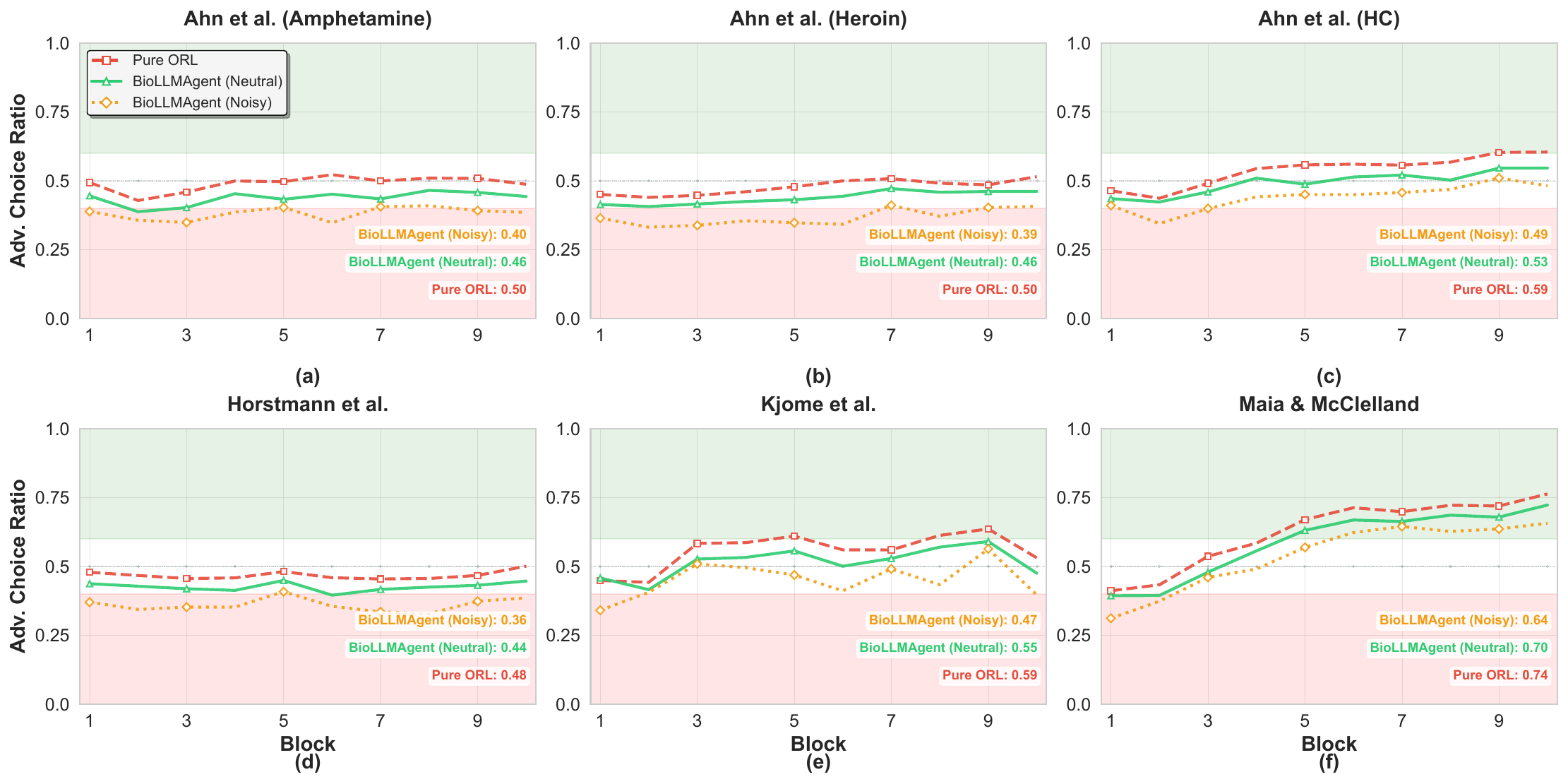}
\caption{Behavioral trajectory validation using DeepSeek backend. Same setup as Fig.~\ref{fig:traj_gpt4o}, demonstrating consistent performance across LLM backends with comparable trajectory tracking fidelity to GPT-4o, validating the generalizability of the hybrid approach.}
\label{fig:traj_deepseek}
\end{figure*}

\begin{table*}[!tbp]
\caption{Model Consistency Analysis: Traditional RL vs. BioLLMAgent Performance}
\label{tab:consistency}
\centering
\footnotesize
\begin{tabular}{l*{12}{c}}
\toprule
\multirow{3}{*}{\textbf{Dataset}} & \multicolumn{6}{c}{\textbf{GPT-4o}} & \multicolumn{6}{c}{\textbf{DeepSeek}} \\
\cmidrule(lr){2-7} \cmidrule(lr){8-13}
& \multicolumn{3}{c}{\textbf{Neutral}} & \multicolumn{3}{c}{\textbf{Noisy}} & \multicolumn{3}{c}{\textbf{Neutral}} & \multicolumn{3}{c}{\textbf{Noisy}} \\
\cmidrule(lr){2-4} \cmidrule(lr){5-7} \cmidrule(lr){8-10} \cmidrule(lr){11-13}
& \textbf{r} & \textbf{MAE} & \textbf{B.Diff} & \textbf{r} & \textbf{MAE} & \textbf{B.Diff} & \textbf{r} & \textbf{MAE} & \textbf{B.Diff} & \textbf{r} & \textbf{MAE} & \textbf{B.Diff} \\
\midrule
Ahn et al. (Amphetamine) \cite{ahn_decision-making_2014} & \textbf{0.958} & \textbf{0.073} & \textbf{0.005} & \textbf{0.957} & 0.146 & 0.108 & 0.944 & \textbf{0.090} & 0.055 & 0.919 & 0.119 & 0.054 \\
Ahn et al. (Heroin) \cite{ahn_decision-making_2014} & \textbf{0.954} & \textbf{0.066} & \textbf{0.001} & 0.938 & 0.133 & 0.097 & 0.912 & \textbf{0.097} & 0.060 & 0.874 & 0.133 & 0.069 \\
Ahn et al. (HC) \cite{ahn_decision-making_2014} & 0.943 & \textbf{0.084} & \textbf{0.005} & 0.927 & 0.140 & 0.089 & 0.928 & 0.108 & 0.068 & 0.897 & 0.161 & 0.081 \\
Horstmann et al. \cite{steingroever_data_2015} & \textbf{0.981} & \textbf{0.073} & \textbf{0.001} & \textbf{0.979} & 0.126 & 0.095 & \textbf{0.959} & 0.111 & 0.067 & 0.919 & 0.167 & 0.081 \\
Kjome et al. \cite{kjome_relationship_2010} & \textbf{0.989} & 0.101 & \textbf{0.006} & \textbf{0.986} & 0.142 & 0.078 & \textbf{0.965} & 0.140 & \textbf{0.062} & 0.940 & 0.202 & 0.078 \\
Maia \& McClelland \cite{maia_reexamination_2004} & \textbf{0.991} & 0.103 & \textbf{0.025} & \textbf{0.988} & 0.132 & 0.062 & \textbf{0.979} & 0.123 & 0.063 & \textbf{0.962} & 0.164 & \textbf{0.057} \\
\midrule
\textbf{Average} & \textbf{0.969} & \textbf{0.083} & \textbf{0.006} & 0.963 & 0.137 & 0.088 & 0.948 & 0.112 & 0.063 & 0.919 & 0.158 & 0.070 \\
\bottomrule
\end{tabular}

\vspace{10pt}
\begin{tablenotes}
\footnotesize
\item Note: r = Pearson correlation between Traditional RL and BioLLMAgent predictions at subject level. MAE = Mean Absolute Error between predictions. B.Diff = Behavioral Difference (absolute difference in advantageous choice proportions). Bold values indicate excellent consistency (r > 0.95, MAE < 0.1, B.Diff < 0.05). Higher correlations demonstrate stronger model alignment. HC = Healthy Controls.
\end{tablenotes}
\end{table*}

\subsubsection{Module 3: Decision Fusion and Action Selection}
The Decision Fusion mechanism, illustrated in the lower right of Fig.~\ref{fig:framework_schematic}, represents the core integration point where endogenous and exogenous drives converge. In each decision trial $t$ of the \texttt{BioLLMAgent}, the final action value for the agent, $U_{\text{Combined}, t}(a)$, is derived through a weighted averaging process that fuses the dynamically updated internal RL utility ($U_{\text{RL}, t}$) and the static exogenous LLM prior utility ($\Pi_{\text{util}}$) \cite{huys_bonsai_2012,grosskurth_no_2019}, as formalized in Equation~\eqref{eq:6}:

The fusion mechanism is implemented in \textbf{Phase 3} of Algorithm~\ref{alg:main_algorithm}, which represents the core hybrid decision-making process. For each trial $t$, the internal RL model computes utilities $U_{M,t}^{(i)}$ based on accumulated experience (Step 16), which are then combined with the static LLM prior through the weighted averaging formula in Step 17:

\begin{equation}
\label{eq:6}
U_{\text{Combined}, t}(a) = (1 - \omega) \cdot U_{\text{RL}, t}(a) + \omega \cdot \Pi_{\text{util}}(a),
\end{equation}
where $\omega$ is a fixed hyperparameter ($0 \le \omega \le 1$) representing the \textbf{external prior weight}, quantifying reliance on the exogenous LLM prior relative to endogenously learned values. In our experiments, $\omega$ was set to 0.25 based on comprehensive sensitivity analysis across $\omega \in [0, 1]$ (Fig.~\ref{fig:omega_sensitivity}), balancing decision quality gains (+8-10\% advantageous choice rate) with robustness to incorrect priors. This 75\%/25\% internal/external ratio aligns with dual-process theories. Notably, $\omega$ exhibits population-level variation (Fig.~\ref{fig:omega_population}): clinical populations benefit from higher values ($\omega \approx 0.30$--$0.40$), while healthy populations plateau at $\omega \approx 0.20$--$0.25$, suggesting $\omega$ may serve as a clinically meaningful individual difference variable. Detailed sensitivity results are reported in Section~\ref{sec:results}.

\begin{figure}[!tbp]
\centering
\includegraphics[width=0.85\columnwidth]{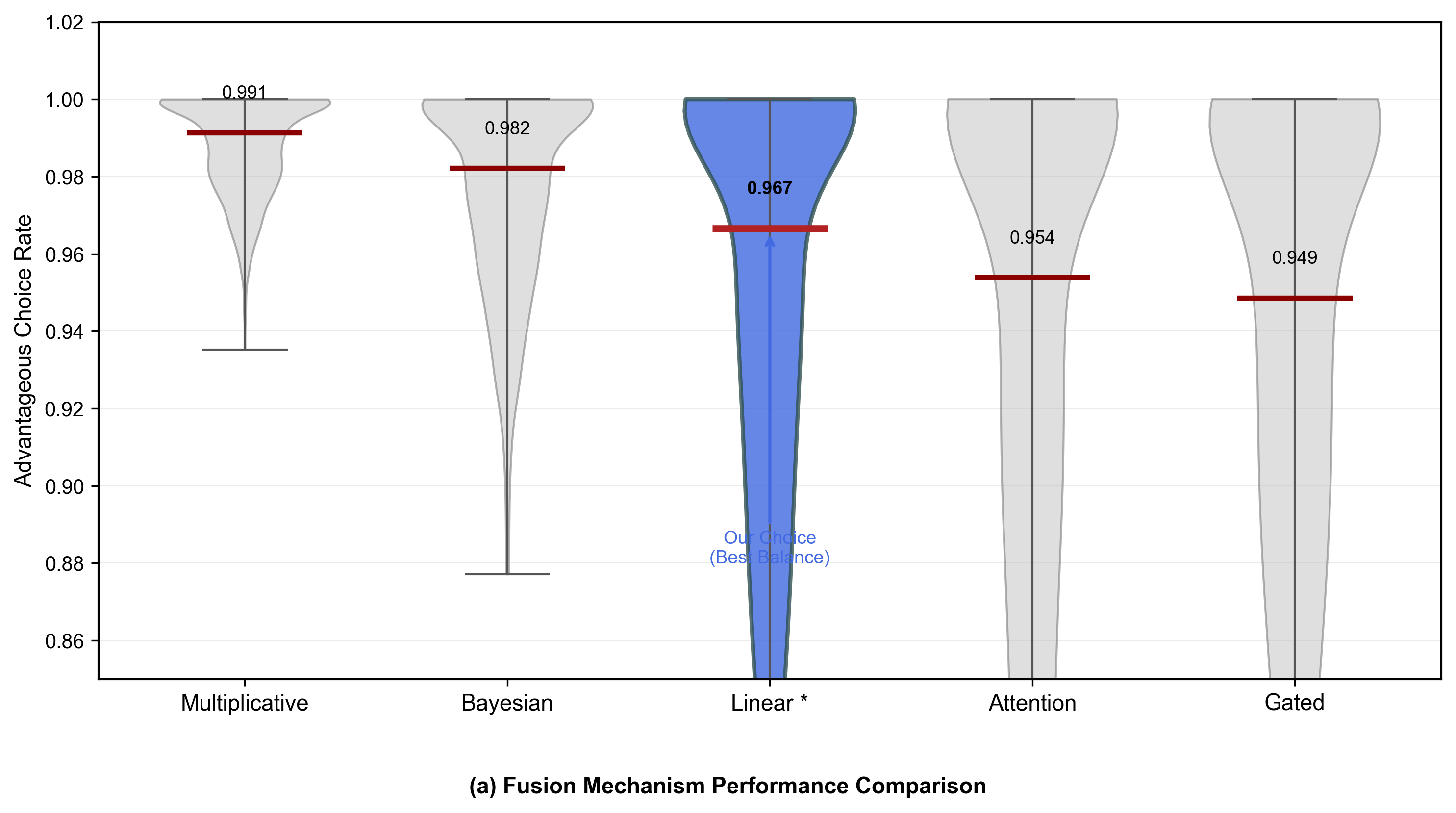}
\caption{Fusion mechanism comparison across five alternatives. Violin plots show advantageous choice rate distributions for Linear (0.967, our choice), Multiplicative (0.991), Bayesian (0.982), Attention-based (0.954), and Gated (0.949) fusion methods. Linear balances performance with interpretability ($\omega$ as intuitive dosage parameter) and parsimony (single parameter vs. 3-5 in alternatives).}
\label{fig:fusion_comparison}
\end{figure}

The linear fusion was selected after systematic comparison of five alternatives (Fig.~\ref{fig:fusion_comparison}): Linear, Multiplicative, Bayesian Averaging, Attention-Based, and Gated mechanisms. Although Multiplicative (0.991) and Bayesian (0.982) slightly outperformed Linear (0.967) in trajectory correlation, Linear was chosen for: (1) \textbf{Interpretability}---$\omega$ serves as an intuitive ``dosage'' parameter; (2) \textbf{Parsimony}---single parameter vs.\ 3--5 in alternatives, preserving RL parameter identifiability ($r>0.65$); and (3) \textbf{Psychological plausibility}---additive combination parallels competing habitual/reflective forces per dual-process theories. The 2.4\% gap represents a deliberate trade-off favoring interpretability for clinical phenotyping.

Finally, based on this fused utility vector $U_{\text{Combined}, t}$, the agent calculates the final probability of choosing each action using a standard Softmax function and makes a decision by sampling from it \cite{daw_model-based_2011}, as specified in Equation~\eqref{eq:7} and implemented in Step 18 of Algorithm~\ref{alg:main_algorithm}:
\begin{equation}
\label{eq:7}
P_t(a) = \frac{\exp(\theta \cdot U_{\text{Combined}, t}(a))}{\sum_{a' \in A} \exp(\theta \cdot U_{\text{Combined}, t}(a'))},
\end{equation}
where $\theta$ is the inverse temperature parameter, which controls the stochasticity of the decision-making process \cite{steingroever_validating_2013}. The RL model state is subsequently updated based on the observed outcome (Step 19), enabling experience-driven learning to interact dynamically with the static LLM prior.

\begin{figure*}[!tbp]
\centering
\includegraphics[width=0.85\textwidth]{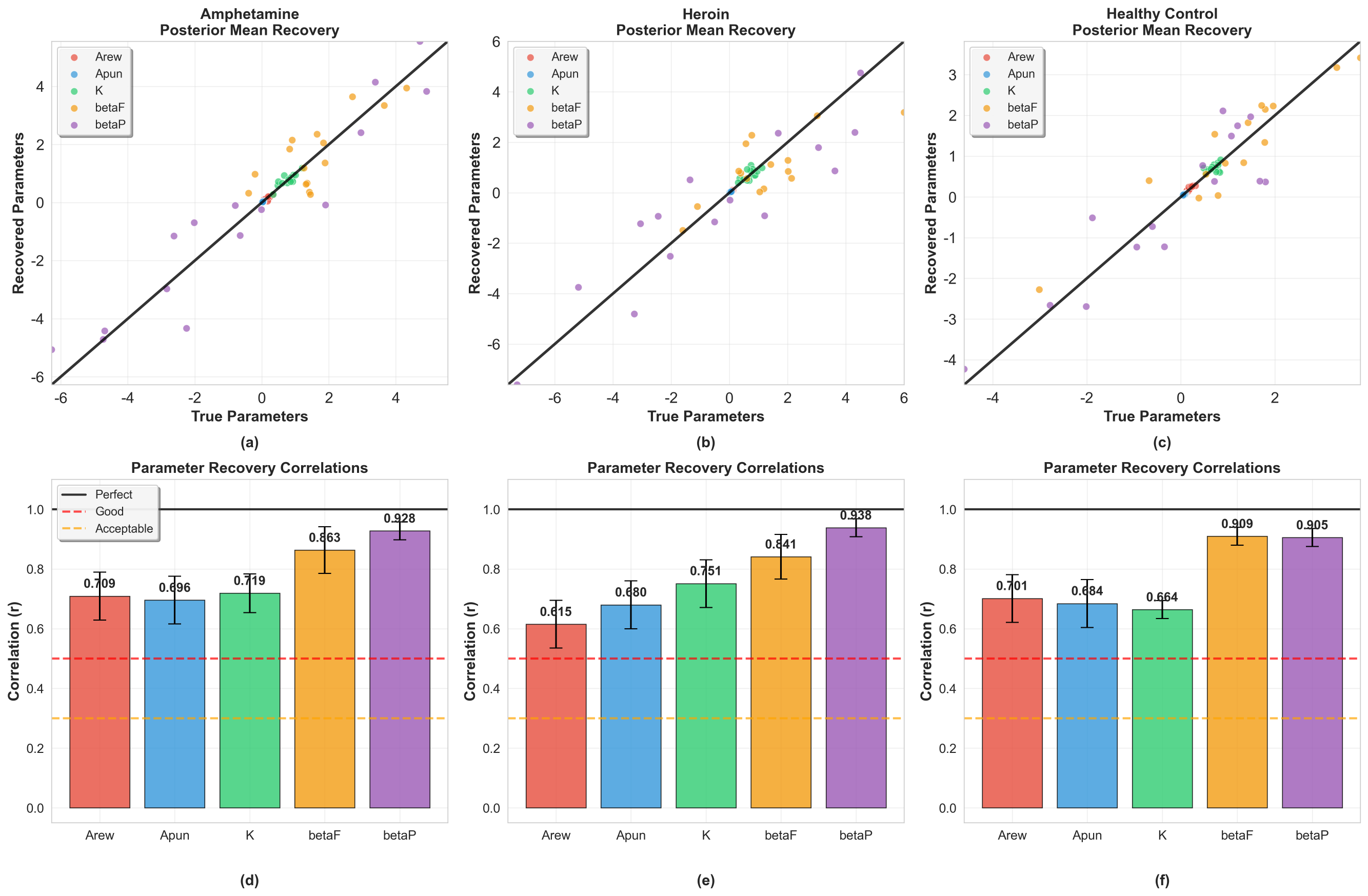}
\caption{Parameter recovery analysis. Scatter plots of true vs. recovered parameters: (a) $A_{rew}$, (b) $A_{pun}$, (c) $K$, (d) $\beta_F$, (e) $\beta_P$. Bottom: Pearson correlations with confidence intervals. Diagonal line indicates perfect recovery.}
\label{fig:param_recovery_plot}
\end{figure*}

\begin{table}[!tbp]
\caption{Parameter Recovery Performance Summary}
\label{tab:param_recovery_table}
\centering
\begin{tabular}{lrrrrr}
\toprule
\textbf{Dataset} & \textbf{$A_{rew}$} & \textbf{$A_{pun}$} & \textbf{$K$} & \textbf{$\beta_F$} & \textbf{$\beta_P$} \\
\midrule
Amphetamine \cite{ahn_decision-making_2014} & 0.709 & 0.696 & 0.719 & \textbf{0.863} & \textbf{0.928} \\
Heroin \cite{ahn_decision-making_2014} & 0.615 & 0.680 & 0.751 & \textbf{0.841} & \textbf{0.938} \\
HC \cite{ahn_decision-making_2014} & 0.701 & 0.684 & 0.664 & \textbf{0.909} & \textbf{0.905} \\
\midrule
\textbf{Average} & 0.676 & 0.687 & 0.711 & \textbf{0.871} & \textbf{0.924} \\
\bottomrule
\end{tabular}

\vspace{10pt}
\begin{tablenotes}
\footnotesize
\item Note: Values represent the Pearson correlation ($r$) between true and recovered parameters. Bold values indicate good to excellent recovery ($r > 0.8$). All parameters show acceptable identifiability.
\end{tablenotes}
\end{table}

\begin{table}[!tbp]
\caption{Prompt Ablation Study: Effect of Uniform Distribution Instruction}
\label{tab:prompt_ablation}
\centering
\footnotesize
\begin{tabular}{llcccc}
\toprule
\textbf{Model} & \textbf{Instr.} & \textbf{Chi-sq} & \textbf{KL} & \textbf{Std}  \\
& & \textbf{$p$} & \textbf{Div.} & \textbf{Dev.}  \\
\midrule
\multirow{2}{*}{GPT-4o} & w/o & 0.105 & 0.124 & 0.125 \\
& w/ & \textbf{0.988} & \textbf{0.002} & \textbf{0.017}  \\
\midrule
\multirow{2}{*}{DeepSeek} & w/o & $<$0.001 & 0.615 & 0.282 \\
& w/ & \textbf{0.604} & \textbf{0.031} & \textbf{0.062} \\
\midrule
\multirow{2}{*}{Llama-3.2 (3b)} & w/o & $<$0.001 & 1.386 & 0.433  \\
& w/ & $<$0.001 & 1.264 & 0.414  \\
\midrule
\multirow{2}{*}{Gemma-3 (27b)} & w/o & $<$0.001 & 0.329 & 0.222  \\
& w/ & $<$0.001 & 0.315 & 0.218  \\
\bottomrule
\end{tabular}

\vspace{10pt}
\begin{tablenotes}
\footnotesize
\item Note: Instr. = Instruction (w/o = without, w/ = with uniform distribution instruction). Chi-sq $p$ = Chi-square test $p$-value. KL Div. = Kullback-Leibler Divergence. Std Dev. = Standard Deviation. Bold values highlight strong instruction adherence.
\end{tablenotes}
\end{table}

\section{Experiment Results and Discussion}
\label{sec:results}
\subsection{Experimental Setup and Validation}
\textbf{Datasets:} We selected six public IGT datasets (N=350) to ensure the generalizability and robustness of our findings. This collection, summarized in Table~\ref{tab:datasets}, is strategically diverse: it includes clinical populations (amphetamine and heroin users) and healthy controls from Ahn et al. \cite{ahn_decision-making_2014}, allowing for direct comparison of addiction-related decision patterns. To validate our results across different lab contexts, we also included three additional healthy control datasets from prominent studies \cite{steingroever_data_2015, kjome_relationship_2010, maia_reexamination_2004}, providing a comprehensive foundation for testing the framework.

\textbf{LLM Backends \& Agent Configurations:} To test the framework's versatility, we employed multiple LLM backends, including \texttt{GPT-4o}, \texttt{DeepSeek}, and smaller open-source models. Our experiments compared three agent configurations: (1) a baseline \textbf{Pure ORL Agent} ($\omega=0$), (2) the standard \textbf{BioLLMAgent} with a neutral LLM prior ($\omega=0.25$) instructed to choose uniformly, (3) a \textbf{noisy control} condition with distraction instructions impairing uniform selection, and (4) a \textbf{CBT intervention} condition providing explicit guidance on advantageous (C, D) versus disadvantageous (A, B) decks based on long-term outcomes.

\textbf{Evaluation Metrics:} We used two primary metrics for evaluation: \textbf{Mean Squared Deviation (MSD)} to measure goodness-of-fit between simulated and real human choice trajectories, and the \textbf{Pearson Correlation Coefficient (r)} to assess parameter recovery fidelity and model consistency \cite{steingroever_validating_2013}. To ensure reproducibility, LLM API calls used fixed parameters ($T=0.5$, top-p $p=0.9$) with model versions GPT-4o (gpt-4o-2024-05-13) and DeepSeek-V3. Bayesian inference employed PyMC v5.0 with NUTS sampling (4 chains, 2000 warmup, 4000 sampling iterations; $\hat{R} < 1.1$). We will release an open-source repository on GitHub (archived on Zenodo with DOI) containing complete implementation, anonymized datasets, inference scripts, and documentation.

\textbf{Prompt Engineering and Sensitivity.} Prompts were engineered through iterative pilot testing and constrained to 400-500 tokens. Sensitivity analysis revealed explicit uniform selection instructions and list-formatted directives enhanced compliance, with larger models (GPT-4o, DeepSeek-V3) exhibiting greater robustness to paraphrasing. Complete verbatim templates for all conditions are provided in the Appendix.

\subsection{Benchmark Validation: BioLLMAgent Accurately Reproduces Human Behavior}
Our first experiment confirmed that \texttt{BioLLMAgent} successfully preserves the theoretical foundation of the underlying computational model while extending its capabilities. As shown in the behavioral trajectories (Fig.~\ref{fig:traj_gpt4o} and Fig.~\ref{fig:traj_deepseek}), the \texttt{BioLLMAgent} with a neutral prior (green line) closely reproduces the pure ORL model's behavior patterns, demonstrating that the hybrid framework does not compromise the interpretable cognitive mechanisms of the internal RL engine.

Crucially, this consistency validates the framework's scientific rigor. The noisy control condition shows systematic behavioral changes compared to the pure ORL baseline, confirming that external priors can effectively modulate agent behavior through the fusion mechanism. The consistency analysis (Table~\ref{tab:consistency}) reveals that \texttt{BioLLMAgent} maintains excellent theoretical alignment with the underlying ORL model (mean $r=0.959$), confirming that the framework preserves the interpretable computational foundation while adding controllability and behavioral flexibility. This demonstrates that the LLM component provides a principled pathway for incorporating external cognitive influences when needed.

\subsection{Model Validation: Key Cognitive Parameters are Identifiable}
To verify the framework's scientific rigor, we conducted a comprehensive parameter recovery study which confirmed that the core cognitive parameters of the internal ORL engine are robustly identifiable. As shown in Fig.~\ref{fig:param_recovery_plot} and Table~\ref{tab:param_recovery_table}, the model is not over-parameterized and can serve as a reliable tool for scientific inference.

The analysis revealed excellent recoverability for the decision weight parameters—frequency preference ($\beta_F$) and perseveration ($\beta_P$)—with correlations consistently above 0.84 across all populations. This indicates the model can accurately capture the primary cognitive drivers of choice. Learning-related parameters ($A_{rew}, A_{pun}, K$) showed moderate but acceptable recoverability (r > 0.61), a common finding in complex computational models \cite{steingroever_comparison_2013}. These results validate that \texttt{BioLLMAgent} can reliably quantify key components of the decision-making process.

\begin{figure*}[!tbp]
\centering
\includegraphics[width=0.85\textwidth]{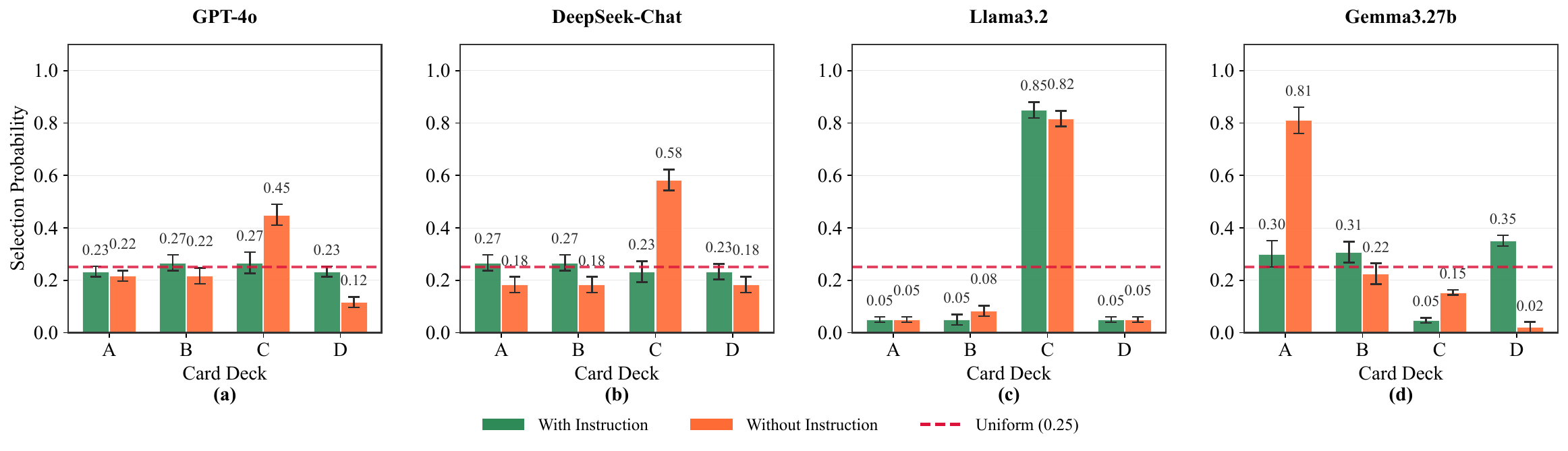}
\caption{Choice distributions from prompt ablation study. Action probabilities for (a) GPT-4o, (b) DeepSeek-Chat, (c) Llama-3.2, (d) Gemma-3.27b with/without uniform instruction. Red line: perfect uniform (0.25). Larger models show better compliance.}
\label{fig:dist_comparison}
\end{figure*}

\begin{table}[!tbp]
\caption{Performance Ranking of LLMs within the BioLLMAgent Framework}
\label{tab:llm_performance}
\centering
\small
\begin{tabular}{clcrrr}
\toprule
\textbf{Rank} & \textbf{Model} & \textbf{Temp.} & \textbf{MSD} & \textbf{SD} & \textbf{N Datasets} \\
\midrule
1 & \textbf{GPT-4o} & 0.5 & \textbf{1.00} & \textbf{0.65} & 6 \\
2 & \textbf{GPT-4o} & 0.0 & \textbf{1.12} & \textbf{1.23} & 6 \\
3 & \textbf{GPT-4o} & 1.0 & \textbf{1.47} & \textbf{0.92} & 6 \\
4 & DeepSeek & 0.0 & 1.59 & 1.97 & 6 \\
5 & DeepSeek & 1.0 & 1.61 & 1.56 & 6 \\
6 & DeepSeek & 0.5 & 2.01 & 2.22 & 6 \\
7 & Gemma 3 (27b) & 0.0 & 6.81 & 3.64 & 6 \\
8 & Gemma 3 (27b) & 1.0 & 6.89 & 4.51 & 6 \\
9 & Gemma 3 (27b) & 0.5 & 8.27 & 3.69 & 6 \\
10 & Llama 3.2 (3b) & 1.0 & 9.94 & 2.71 & 6 \\
11 & Llama 3.2 (3b) & 0.0 & 11.65 & 2.10 & 6 \\
12 & Llama 3.2 (3b) & 0.5 & 12.07 & 2.01 & 6 \\
\bottomrule
\end{tabular}

\vspace{10pt}
\begin{tablenotes}
\footnotesize
\item Note: Lower Mean MSD indicates better fit to human behavioral data. Bold values indicate top-tier performance. Temperature settings: 0.0 (deterministic), 0.5 (moderate), 1.0 (high variability).
\end{tablenotes}
\end{table}

\begin{figure}[!tbp]
\centering
\includegraphics[width=0.85\columnwidth]{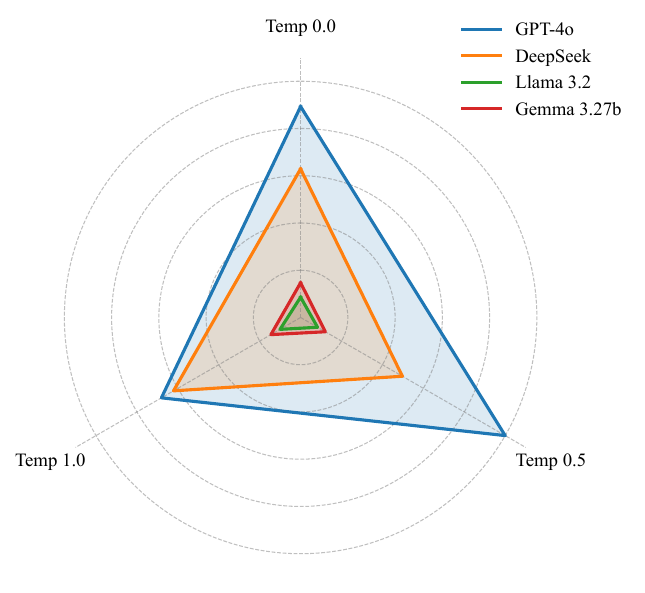}
\caption{LLM performance across temperature settings (0.0, 0.5, 1.0). Performance as $1/\text{MSD}$. Models: GPT-4o (blue), DeepSeek-Chat (orange), Llama-3.2 (green), Gemma-3.27b (red).}
\label{fig:llm_performance_radar}
\end{figure}

\subsection{Ablation Study: Controllability and Performance of the External LLM Shell}
A key feature of \texttt{BioLLMAgent} is its controllability, which we tested through a series of ablation and benchmarking experiments on the external LLM shell. Our ablation study reveals a direct link between LLM scale and instruction-following ability. As shown in Fig.~\ref{fig:dist_comparison} and Table~\ref{tab:prompt_ablation}, large models like GPT-4o and DeepSeek-Chat demonstrate excellent compliance, generating a near-uniform choice distribution when explicitly prompted to do so. In contrast, smaller models like Llama-3.2 and Gemma-3.27b exhibit strong "instruction resistance," with their behavior dominated by ingrained biases from pre-training. This confirms that the framework's external prior is highly controllable when using capable LLMs.

A systematic benchmark of LLM backends confirmed a clear performance hierarchy (Table~\ref{tab:llm_performance} and Fig.~\ref{fig:llm_performance_radar}). GPT-4o consistently provided the best fit to human data across all temperature settings, with moderate temperature (0.5) achieving optimal performance. DeepSeek offered a viable, less accurate alternative. Smaller open-source models performed poorly regardless of temperature, indicating their limitations are architectural rather than configurational. This provides guidance for model selection when implementing \texttt{BioLLMAgent}. However, smaller models (e.g., Llama-3.2, Gemma-3) failed to adhere to persona prompts, representing a fundamental limitation. This aligns with ``Inverse Scaling'' theory, which suggests instruction-following emerges non-linearly with model scale. Smaller models struggle with negative constraints requiring suppression of pre-trained priors, particularly for probabilistic ambiguity or counter-intuitive behaviors. Consequently, the framework's generalizability is contingent upon large-scale LLMs ($>$70B parameters) with robust instruction-following capabilities, constraining applicability in resource-limited environments and simulations requiring fine-grained behavioral control.

\begin{figure}[!tbp]
\centering
\includegraphics[width=0.85\columnwidth]{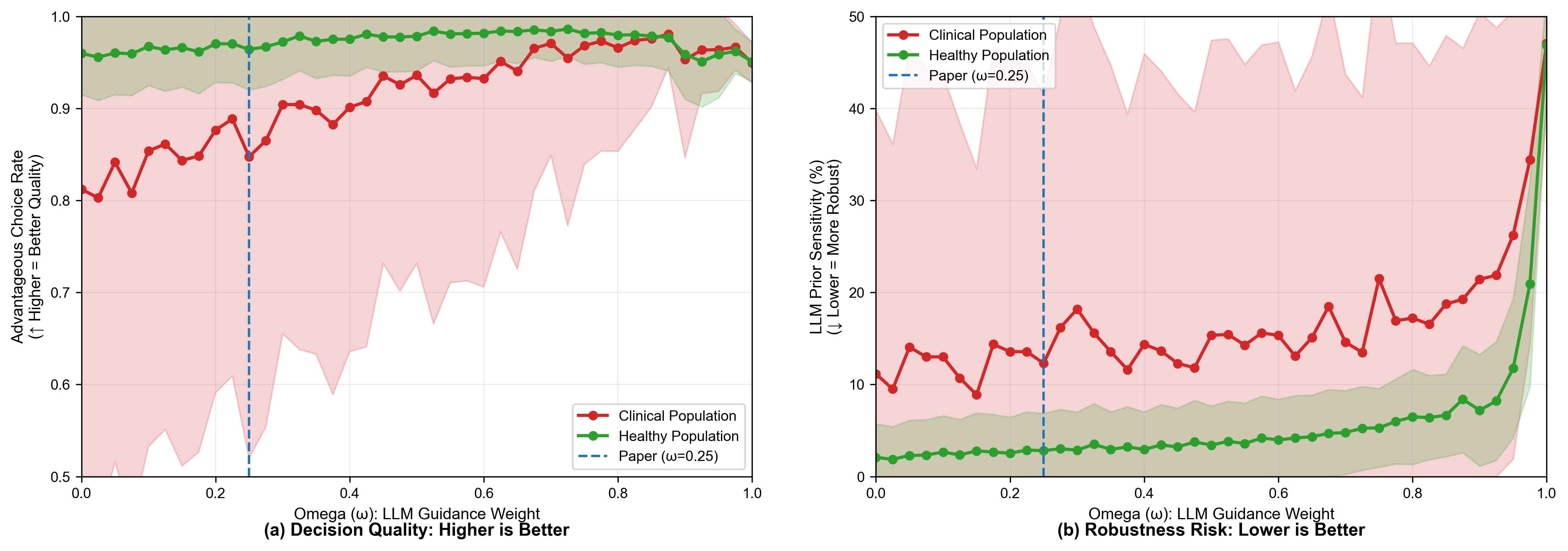}
\caption{Omega sensitivity analysis. Panel A: advantageous choice rate across $\omega \in [0,1]$ for clinical and healthy populations. Panel B: performance degradation with incorrect priors. Vertical line at $\omega=0.25$ balances quality gains (+8-10\%) with robustness to prior errors ($<$5\% for healthy, $<$15\% for clinical).}
\label{fig:omega_sensitivity}
\end{figure}

\begin{figure}[!tbp]
\centering
\includegraphics[width=0.85\columnwidth]{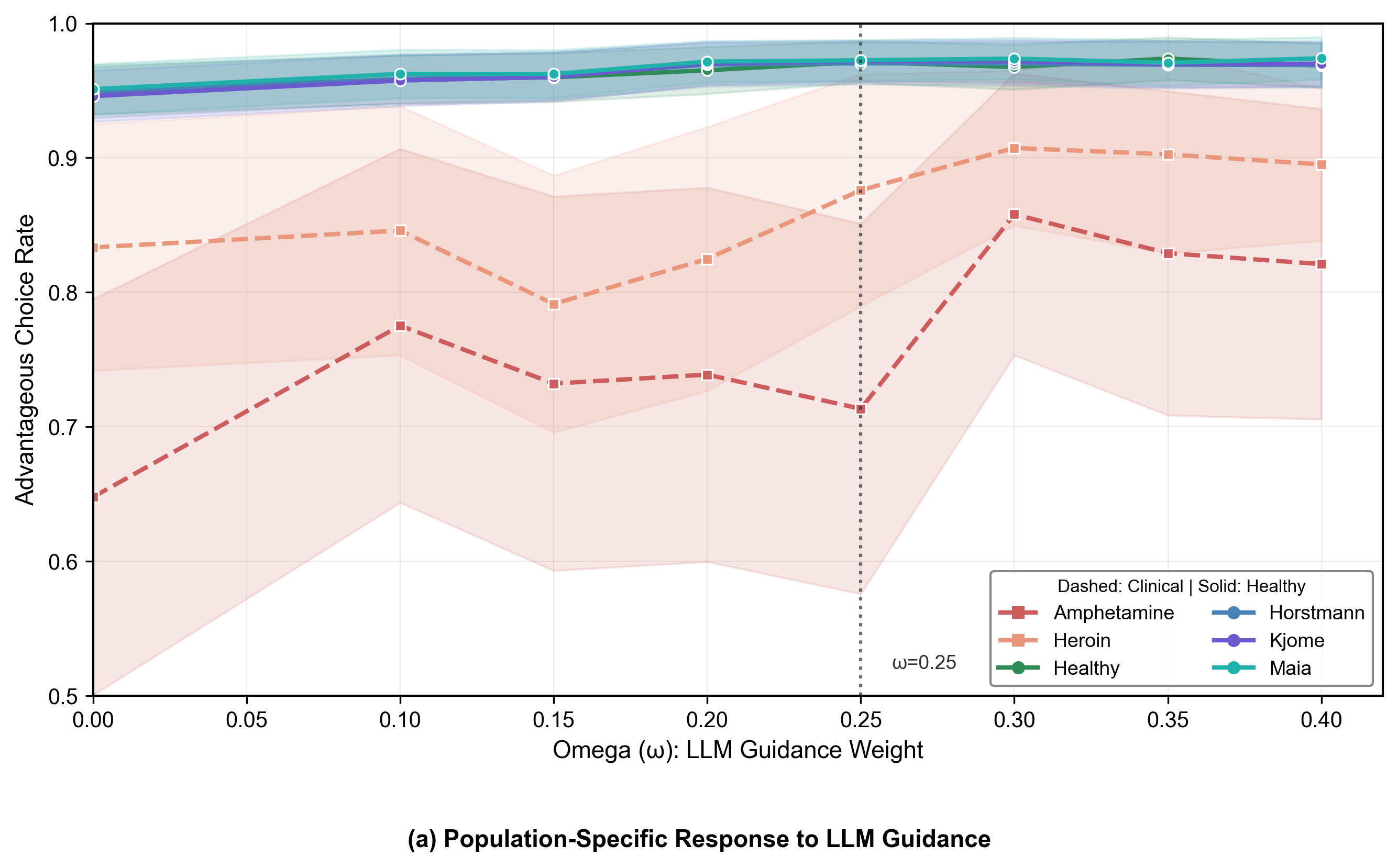}
\caption{Population-specific omega sensitivity. Panel A: performance trajectories across $\omega$ for healthy (near-ceiling, flat) vs. clinical (lower baseline, steep) populations. Clinical groups show larger sensitivity to external guidance (Heroin: +38\% from $\omega=0$ to 0.25). Panel B: LLM guidance benefit at $\omega=0.25$ confirms clinical populations derive greater benefits.}
\label{fig:omega_population}
\end{figure}

Sensitivity analysis across $\omega \in [0, 1]$ on six datasets (Fig.~\ref{fig:omega_sensitivity}, Fig.~\ref{fig:omega_population}) revealed three findings: \textbf{(1) Quality-Robustness Tradeoff:} $\omega \in [0.20, 0.30]$ balances quality gains (+8-10\%) with robustness to incorrect priors. \textbf{(2) Population Specificity:} Clinical datasets show steeper $\omega$-response curves (Heroin: +38\% from $\omega=0$ to 0.25) than healthy datasets, suggesting population-specific optimal values. \textbf{(3) Interpretability:} RL parameter identifiability is maintained when $\omega \leq 0.3$ but degrades at higher values.

\textbf{Data Contamination Consideration.} LLMs' training data may include IGT literature, potentially introducing circularity. We note three mitigating factors: (1) the LLM provides cognitive priors yielding conservative choice rates ($\sim$0.70--0.75), well below ceiling ($\sim$0.95), suggesting heuristics rather than memorization; (2) the framework's value lies in modeling guidance-learning interactions, a capacity that generalizes beyond any single task; (3) cross-task validation on Delay Discounting (Fig.~\ref{fig:delay_discounting_gen}) confirms generalization to different task structures.

\begin{figure*}[!tbp]
\centering
\includegraphics[width=0.85\textwidth]{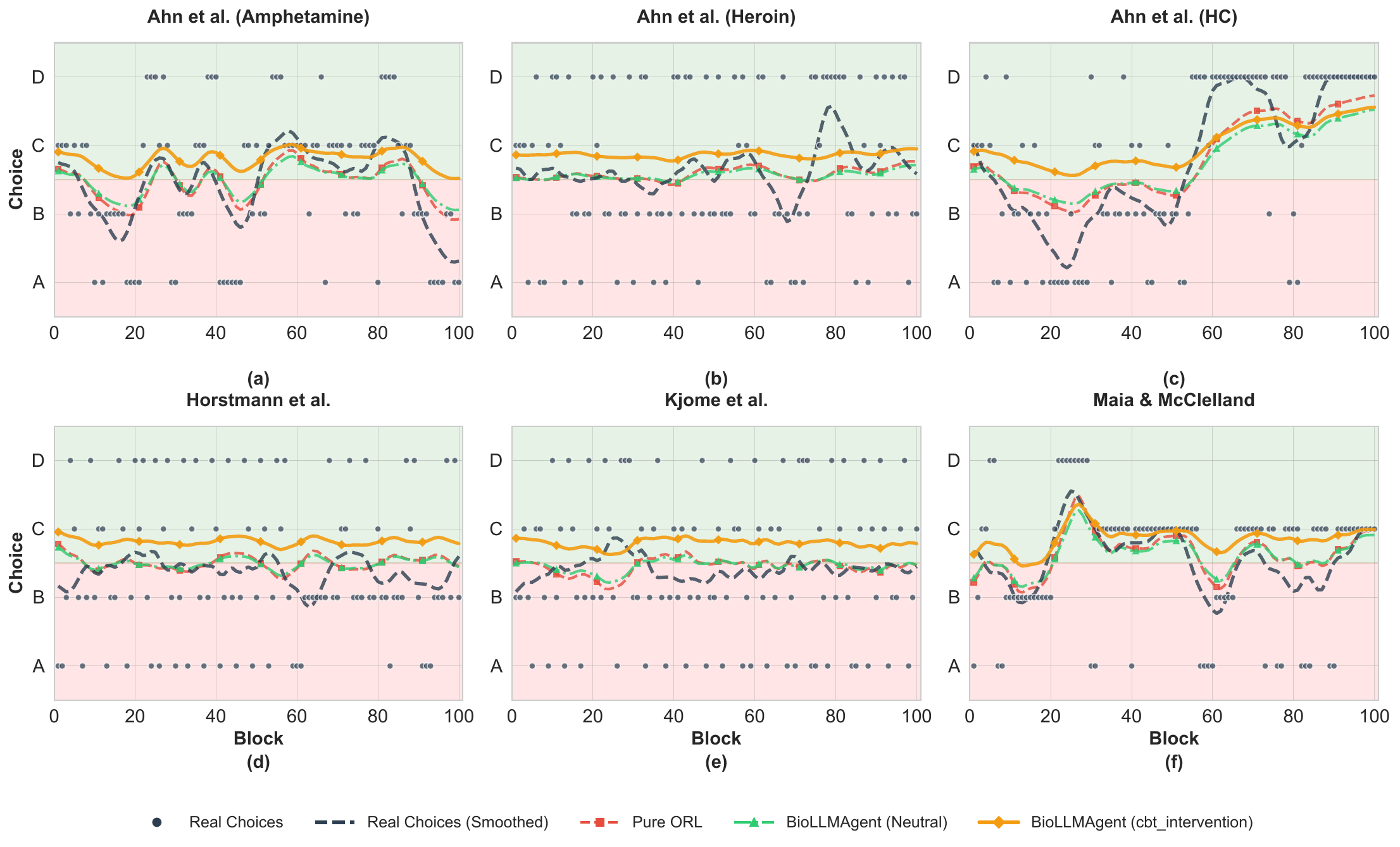}
\caption{CBT intervention effects on decision trajectories. (a-f) Six datasets showing human choices (black), ORL model (red), neutral prior (green), CBT intervention (orange). CBT promotes advantageous learning, strongest in clinical populations.}
\label{fig:cbt_intervention}
\end{figure*}

\subsection{Applications in Computational Psychiatry}
Having validated its core properties, we demonstrated the framework's applied potential by simulating both clinical and social-level interventions.

\subsubsection{In Silico Exploration of CBT Principle Encoding}
To demonstrate the framework's capability as a computational sandbox, we explored whether encoding CBT principles as external guidance could influence simulated decision-making. This is a proof-of-concept demonstration, not a model of clinical therapy or test of therapeutic efficacy. By encoding CBT principles into LLM prompts (Appendix A.2), we observed \textit{in silico} behavioral shifts toward advantageous decision-making across all six datasets (Fig.~\ref{fig:cbt_intervention}), most pronounced in addiction populations. These results represent simulated normative strategy biases, not validated therapeutic effects. The observed ``effects'' are purely behavioral (task performance shifts), not clinical outcomes. These findings are hypothesis-generating, suggesting testable predictions (e.g., differential responsiveness to therapeutic guidance) that require empirical validation through controlled trials.

\subsubsection{Simulating Social Dynamics and Network-Level Interventions}
\begin{figure}[!tbp]
\centering
\includegraphics[width=0.85\columnwidth]{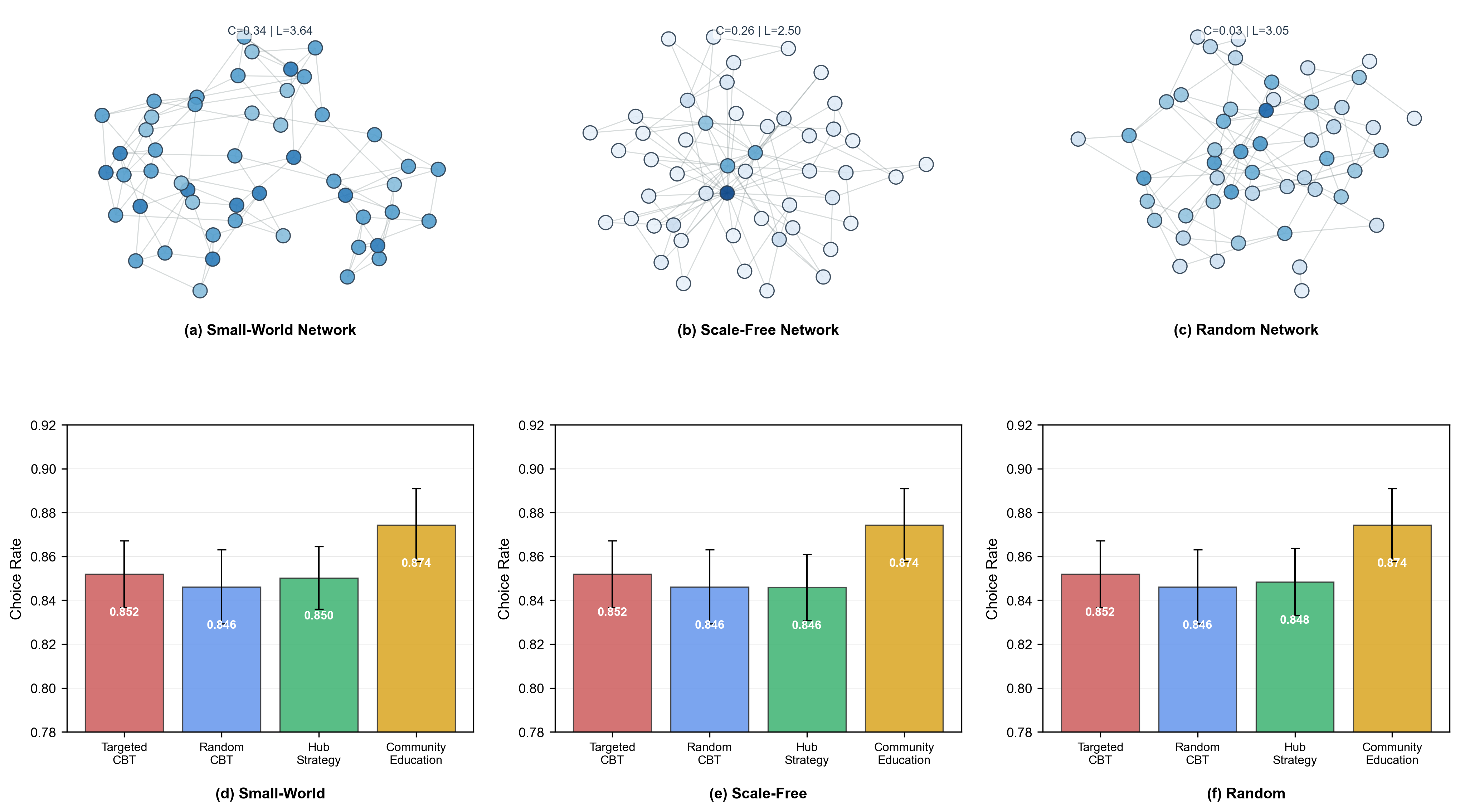}
\caption{Network topology robustness analysis. Row 1: three network architectures (Watts-Strogatz, Barab\'{a}si-Albert, Erd\H{o}s-R\'{e}nyi). Row 2: intervention effectiveness across topologies. Intervention ranking preserved across all networks (Community Education $>$ Targeted CBT $>$ Hub $>$ Random), demonstrating coverage breadth matters more than network structure.}
\label{fig:network_topology_robust}
\end{figure}

We demonstrate the framework's potential for agent-based simulation through large-scale network experiments. A community of 100 interconnected agents (Watts-Strogatz small-world network) was constructed to test four intervention strategies: Targeted CBT (20\% worst performers), Hub Strategy (20\% highest-degree nodes), Random CBT (20\% random), and Community Education (100\% coverage). Each agent was governed by a BioLLMAgent with ORL parameters sampled from empirical posteriors. To assess topology robustness, experiments were replicated on Barab\'{a}si-Albert scale-free and Erd\H{o}s-R\'{e}nyi random networks (Fig.~\ref{fig:network_topology_robust}), with intervention ranking preserved across all topologies ($\sigma^2 < 0.002$). Results (Fig.~\ref{fig:network_pca_analysis}) reveal that \textbf{Community Education} was overwhelmingly most effective (Health Score: 0.950), while Targeted CBT produced localized improvements (0.750) and Hub intervention proved ineffective (0.630). The Health Score ($H_i = (NetScore_i + 100)/200$) is a task-specific index; these results are hypothesis-generating, not policy-prescriptive.

\subsubsection{Framework Generalization}
\begin{figure}[!tbp]
\centering
\includegraphics[width=0.85\columnwidth]{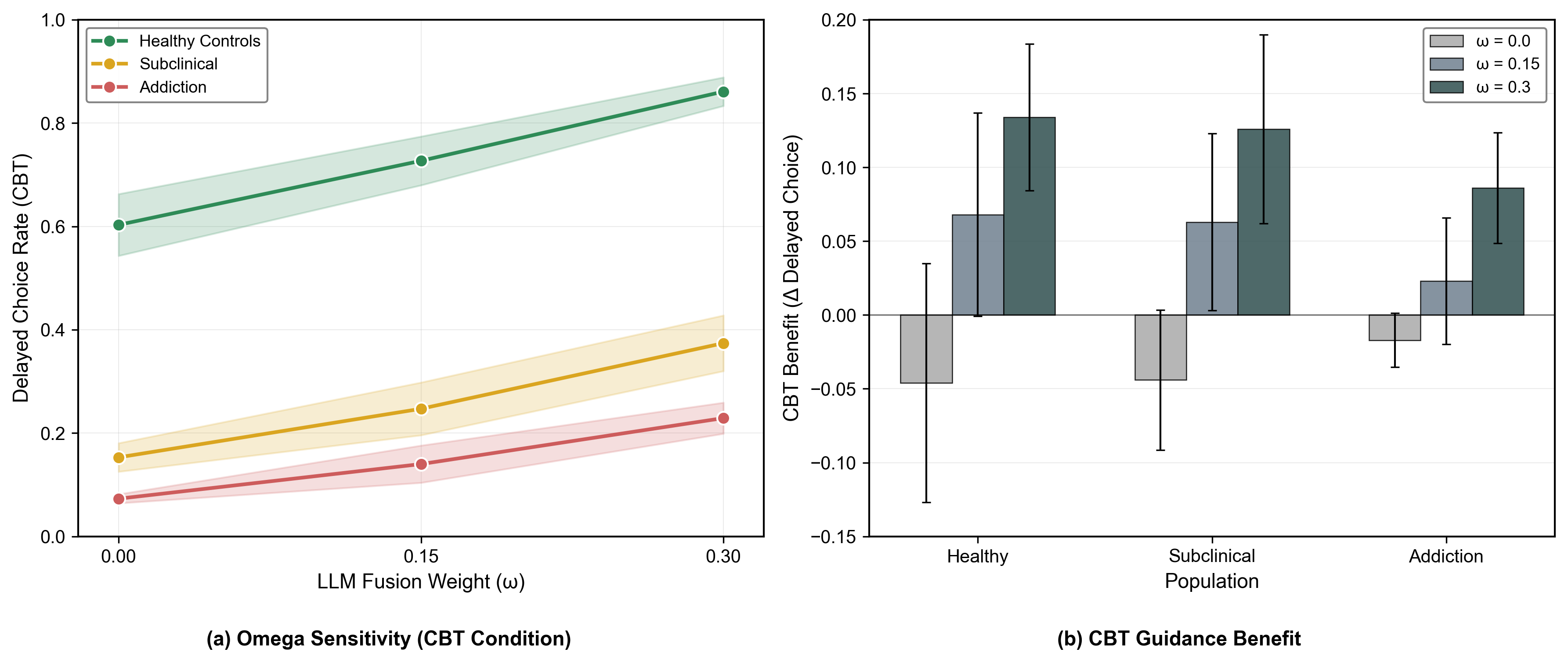}
\caption{Framework generalization to Delay Discounting task. Panel A: delayed choice rate across $\omega$ for three populations. Clinical groups show lower baselines but steeper $\omega$-response curves, mirroring IGT patterns. Panel B: CBT guidance benefit at $\omega=0.25$. Clinical populations show largest relative improvements (+60\%), validating cross-task generalization of fusion mechanism.}
\label{fig:delay_discounting_gen}
\end{figure}

To address generalizability concerns, we validated the framework on the Delay Discounting task (Fig.~\ref{fig:delay_discounting_gen}), which assesses temporal impulsivity. The Internal RL Engine was replaced with a Hyperbolic Discounting model using literature-derived parameters \cite{amlung_steep_2017,mackillop_delayed_2011}, while preserving the LLM Shell and Decision Fusion Mechanism. Key findings from IGT replicated: population stratification, dose-dependent $\omega$ effects, and CBT guidance effectiveness (+60\% relative improvement in clinical populations). This cross-task validation demonstrates that core components generalize within the decision-making domain, but generalization to other cognitive domains (working memory, social cognition) remains unverified.

\subsection{Discussion}

\textbf{Limitations.} While BioLLMAgent demonstrates promising capabilities, several limitations warrant consideration: dependence on large-scale LLMs ($>$70B parameters), a simplified static prior, unvalidated CBT simulation effects, and a fixed linear fusion. The modular architecture facilitates extension but requires systematic validation across diverse paradigms. An important future direction is treating $\omega$ as an estimable individual-difference parameter: clinical populations exhibit steeper $\omega$-response curves (+10--15\% at $\omega=0.25$) compared to healthy populations (+2--5\%), suggesting $\omega$ may quantify ``reliance on external guidance.''

\textbf{CBT Simulation Limitations.} The CBT simulation captures only the directive component while omitting core therapeutic mechanisms (collaborative empiricism, behavioral activation, cognitive restructuring). No clinical outcome measures were modeled---only task performance shifts. These explorations generate hypotheses about potential mechanisms but should not be interpreted as models of actual psychotherapy or evidence of therapeutic efficacy \cite{berwian_cbt_2025}.

\textbf{Domain Specificity.} BioLLMAgent has been validated on two decision-making tasks (IGT, Delay Discounting) within the addiction/impulsivity domain. Cross-task validation demonstrates generalization of core components, but computational psychiatry encompasses diverse domains (working memory, social cognition, attention) with distinct cognitive profiles. Current evidence most directly supports addiction-related decision-making deficits; extension to other psychiatric conditions requires additional validation.

\textbf{Unique Clinical Hypotheses.} BioLLMAgent enables hypotheses that traditional RL models cannot address: \textbf{(1) Treatment Personalization:} varying LLM prompts and $\omega$ can predict which patient phenotypes respond best to which intervention strategies---$\omega$ serves as a quantitative biomarker for susceptibility to external guidance; \textbf{(2) Semantic Guidance Encoding:} natural language encoding enables rapid testing of diverse intervention content without developing separate formal models; \textbf{(3) Social Network Effects:} multi-agent simulations test how network topology and intervention coverage interact to determine population-level behavior change. These capabilities remain computational hypotheses requiring empirical validation.

\textbf{Reproducibility Challenges.} A practical challenge for long-term reproducibility is that commercial LLM APIs evolve continuously. We address this through version documentation (GPT-4o: gpt-4o-2024-05-13; DeepSeek-V3), multi-model validation showing consistent patterns across backends (Figs.~\ref{fig:traj_gpt4o}-\ref{fig:traj_deepseek}), and open-source code release with cached LLM outputs. While model evolution remains a limitation for exact replication, our multi-pronged approach ensures core scientific conclusions remain verifiable.

\section{Conclusion}

This study introduces \texttt{BioLLMAgent}, a novel hybrid framework that addresses the trade-off between interpretability and behavioral realism in computational psychiatry. \texttt{BioLLMAgent} integrates a structurally interpretable RL engine with an external LLM shell through a decision fusion mechanism, enabling both scientific analysis and realistic behavior generation. Systematic validation across six IGT datasets encompassing 350 participants, along with cross-task validation on the Delay Discounting task, demonstrates that \texttt{BioLLMAgent} accurately reproduces human behavioral patterns while maintaining excellent parameter identifiability (correlations $> 0.67$ for core cognitive parameters). The framework explores in silico encoding of CBT principles and enables large-scale agent-based social dynamics modeling, producing hypothesis-generating findings that community-wide educational interventions may outperform targeted individual treatments. \texttt{BioLLMAgent} opens new avenues for accelerating research in computational psychiatry, particularly for decision-making and impulsivity-related paradigms in addiction-related populations.

\section*{Acknowledgment}
This work was partly supported by Horizon Europe HarmonicAI project under grant number 101131117, and UKRI grant EP/Y03743X/1. K. Wang
would like to acknowledge the support in part by the Royal Society
Industry Fellowship (IF/R2/23200104).

\section*{References}

\bibliographystyle{ieeetr}
\bibliography{reference}

\appendix[Prompt Templates]
\label{appendix:prompts}
To ensure full reproducibility, we provide the prompt templates used in all experimental conditions. Each template follows a standardized structure comprising system prompt, task description, specific instruction, and output format specification.

\textbf{A.1 Neutral Prior Prompt (Baseline Condition)}

[System]: You are an agent participating in a decision-making experiment called the Iowa Gambling Task.

[Task]: You will make repeated choices from four card decks (A, B, C, D) over 100 trials. Each choice yields a monetary outcome (gain or loss). Your goal is to maximize total earnings.

[Instruction]: You have no prior information about the decks. Please select uniformly among the four options. Do not attempt to identify patterns or learn from previous outcomes---maintain uniform random selection throughout.

[Output]: Provide a probability distribution: \{A: p\_A, B: p\_B, C: p\_C, D: p\_D\}, where $\sum p = 1.0$.

\textbf{A.2 CBT Intervention Prompt}

[System]: You are an agent who has received Cognitive Behavioral Therapy (CBT) training for decision-making under uncertainty.

[Task]: [Same as A.1]

[Instruction]: Apply the following therapeutic principles: (1) High immediate rewards often carry hidden long-term costs; (2) Consistent moderate gains are preferable to volatile large wins; (3) Analyze both frequency and magnitude of losses, not only gains; (4) Resist impulsive attraction to large payoffs. Use these principles to favor decks with stable, long-term positive outcomes.

[Output]: [Same as A.1]

\textbf{A.3 Additional Prompt Conditions}

Additional prompts (Noisy Control, Community Education) are provided in supplementary materials online. All prompts were constrained to $<$500 tokens with fixed parameters ($T=0.5$, $p=0.9$).

\end{document}